\documentclass[10pt,twocolumn,letterpaper]{article}

\usepackage[pagenumbers]{cvpr} % To force page numbers, e.g. for an arXiv version

\definecolor{cvprblue}{rgb}{0.21,0.49,0.74}
\usepackage[pagebackref,breaklinks,colorlinks,allcolors=cvprblue]{hyperref}
\usepackage[most]{tcolorbox}
\usepackage{booktabs,multirow}
\usepackage{algorithm}
\usepackage{algorithmic}
\usepackage{siunitx}
\usepackage{graphicx}
\usepackage{subcaption}

\usepackage{makecell}
\newcommand{\thU}[2]{\multicolumn{1}{c}{\makecell[c]{\textbf{#1}\\(#2)}}}
\def \one {\mathbf{1}}

\usepackage{amsmath,amssymb,bm}

\newtheorem{theorem}{Theorem}[section]

\def\paperID{1107} % *** Enter the Paper ID here
\def\confName{CVPR}
\def\confYear{2026}

\title{Sparse by Rule: Probability-Based N:M Pruning for Spiking Neural Networks} %\\ with \(\mathcal{O}(M)\) Complexity

\author{Shuhan Ye \and Yi Yu \and Qixin Zhang \and Chenqi Kong \and Qiangqiang Wu \and Xudong Jiang \and Dacheng Tao\\
Nanyang Technological University\\
{\tt\small SHUHAN006@e.ntu.edu.sg}
}

\begin{document}
\maketitle
\begin{abstract}
Brain-inspired Spiking neural networks (SNNs) promise energy-efficient intelligence via event-driven, sparse computation, but deeper architectures inflate parameters and computational cost, hindering their edge deployment. 
Recent progress in SNN pruning helps alleviate this burden, yet existing efforts fall into only two families: \emph{unstructured} pruning, which attains high sparsity but is difficult to accelerate on general hardware, and \emph{structured} pruning, which eases deployment but lack flexibility and often degrades accuracy at matched sparsity. 
In this work, we introduce \textbf{SpikeNM}, the first SNN-oriented \emph{semi-structured} \(N{:}M\) pruning framework that learns sparse SNNs \emph{from scratch}, enforcing \emph{at most \(N\)} non-zeros per \(M\)-weight block. 
To avoid the combinatorial space complexity \(\sum_{k=1}^{N}\binom{M}{k}\) growing exponentially with \(M\), SpikeNM adopts an \(M\)-way basis-logit parameterization with a differentiable top-\(k\) sampler, \emph{linearizing} per-block complexity to \(\mathcal O(M)\) and enabling more aggressive sparsification.
Further inspired by neuroscience, we propose \emph{eligibility-inspired distillation} (EID), which converts temporally accumulated credits into block-wise soft targets to align mask probabilities with spiking dynamics, reducing sampling variance and stabilizing search under high sparsity. 
Experiments show that at \(2{:}4\) sparsity, SpikeNM maintains and even with gains across main-stream datasets, while yielding hardware-amenable patterns that complement intrinsic spike sparsity.
\end{abstract}

\begin{figure*}[!t]
    \centering
    \includegraphics[width=\textwidth]{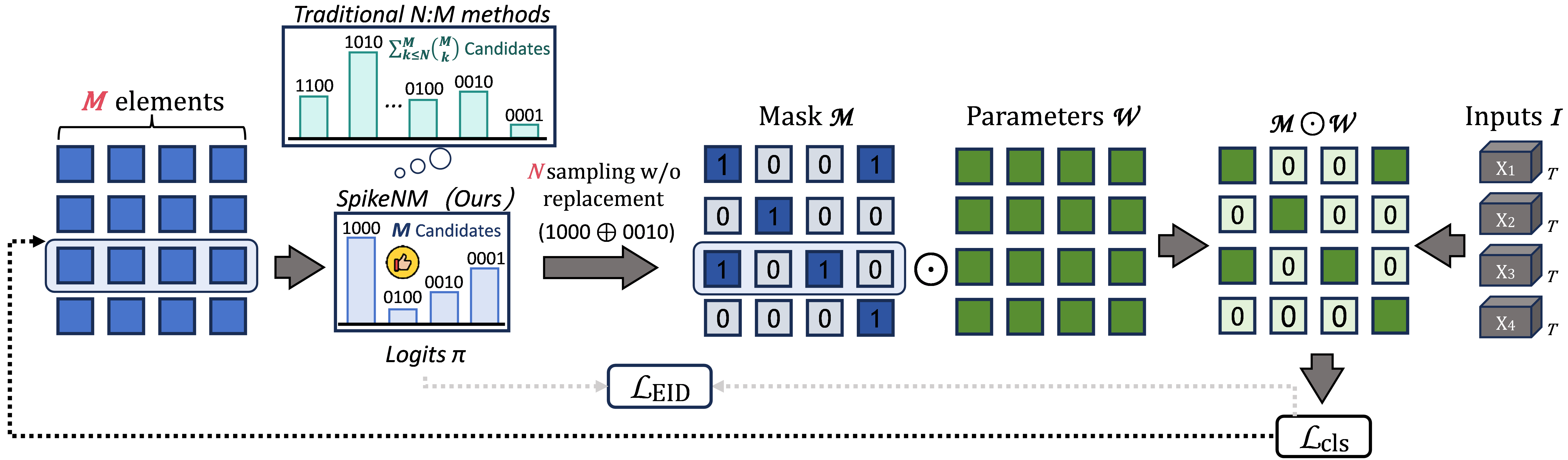}
    \vspace{-6mm}\caption{\textbf{The overview of SpikeNM.} For each block of $M$ weights, we learn logits $\boldsymbol{\pi}$ and \emph{sample $N$ one-hot basis without replacement}. These operations yield an \emph{at-most-$N$} binary mask $\mathcal{M}$, avoiding enumeration of all $\sum_{k\le N}\binom{M}{k}$ candidates used in traditional $N{:}M$ methods (For ease of illustration, we use 2:4 sparsity in the figure). 
    The mask gates parameters $\mathcal{W}$ %($\mathcal{M}!\odot!\mathcal{W}$) 
    during training on inputs of $T$ time steps, optimized by the classification loss $\mathcal{L}_{\text{cls}}$ together with the eligibility-inspired regularizer $\mathcal{L}_{\text{EID}}$. At test time, the learned discrete masks are fixed, producing $N{:}M$ semi-structured sparsity.}
    \label{main}
    \vspace{-5mm}
\end{figure*}

\section{Introduction}
% SNN的优点,但越来越深的架构也不利于边缘部署。
Brain-inspired Spiking neural networks (SNNs), regarded as the third generation of neural networks~\citep{m:97}, have attracted growing interest for their biological plausibility, event-driven computation, and energy efficiency~\citep{subbulakshmi2021biomimetic, hu2023fast}. 
Unlike artificial neural networks (ANNs) that are continuously active, SNNs maintain a temporal state that integrates inputs and fires sparse spikes when crossing a threshold \citep{LIF, zhou2023computational, hu2021spiking}.
These characteristics significantly reduce computational complexity and minimize unnecessary overhead \citep{Davies2018loihi} when deployed on neuromorphic chips \citep{ma2024darwin3}, thus positioning SNNs as an energy-efficient alternative to conventional ANNs.
Despite these efficiency benefits, SNNs still lag behind their ANN counterparts in performance, motivating efforts to increase network depth to close the gap \citep{fang2021deep,highperformance,dldsnn}.
This pursuit imposes substantial costs and undermines edge deployment, as over-parameterization during training and inference inflates computational and memory footprints, motivating solutions that preserve SNN efficiency while curbing parameter and compute growth~\citep{UPR, ckd, li2024towards}.

% SNN剪枝的工作，引出N：M剪枝
SNN pruning has proven to be a viable solution. Recent methods generally fall into two families: \emph{structured} approaches \citep{meng2023efficient,li2024efficient,li2024towards} and \emph{unstructured} approaches \citep{GradR,gou2025dynamic,UPR,shenimproving}.
Unstructured pruning removes individual weights using magnitude or saliency-based criteria, setting less important parameters to zero during training. 
These methods can achieve high sparsity with minimal architectural change, but their \emph{irregular patterns} are difficult to exploit for speedups on general hardware and often require bespoke sparse kernels.
Structured pruning removes channels, filters, or layer-wise neurons, yielding \emph{hardware-friendly} dense tensor shapes that reduce storage requirements and simplify deployment.
However, its coarser granularity often causes greater performance degradation at the same sparsity level compared to unstructured pruning, typically necessitating enormous retraining. 
To reconcile these trade-offs, the semi-structured pruning method was proposed \citep{zhou2021learning, lu2023step, maskllm} for ANNs. 
This method is positioned between structured and unstructured methods, introducing \(N{:}M\) sparsity that leaves only \(N\) non-zero values in each group of \(M\) consecutive weights. 
This regularity has shown tangible acceleration on \emph{Nvidia A100} GPUs~\citep{mishra2021accelerating}, offering the advantage of harmonizing the acceleration benefits of structured patterns and the flexibility of fine-grained sparsity.
In SNNs, this promises to simultaneously exploit intrinsic spike sparsity for unstructured activation savings and \(N{:}M\) weight regularity for hardware-friendly acceleration, yet it has not been explored.

% 我们的方法, 同时SNN中的稀疏出发机制，剪枝比例往往可以达到很高，为了应对搜索空间随M指数增长带来的问题
In this work, we explore and propose \textbf{SpikeNM}, the first SNN-oriented semi-structured pruning method that learns \(N{:}M\) pruned sparse SNNs \emph{from scratch}. 
SpikeNM enforces \emph{at most \(N\) non-zeros per \(M\)-weight block} and is tailored to SNNs. 
Unlike ANNs, the sparse, event-driven nature of SNNs allows them to tolerate more aggressive pruning, yet under such regimes the \(N{:}M\) mask space \(\sum_{k=1}^{N}\binom{M}{k}\) grows combinatorially with \(M\), inflating memory and search variance and impeding scalable training. 
SpikeNM addresses this by replacing subset learning with an \emph{\(M\)-way basis-logit parameterization} and a \emph{differentiable top-\(k\)} sampler: each block maintains \(M\) logits and selects \(N\) indices, retaining the same candidate search space while learning with only \(\mathcal O(M)\) parameters and \(\mathcal O(M)\) memory per block, thus preserving feasibility and scalability under high sparsity.
Beyond this, we encourage \emph{intra-block competition} to retain the most informative connections. 
Inspired by eligibility traces in neuroscience \citep{fremaux2016, singh1996reinforcement}, we introduce an eligibility-inspired distillation (EID) regularizer: we accumulate per-synapse temporal credit into block-wise soft targets and distill them into the mask logits. 
This aligns mask selection with spiking dynamics, reduces sampling variance under sparse and nonstationary activations, and stabilizes convergence at extreme sparsity.

Our main contributions are summarized as follows:
\begin{itemize}
\item \textbf{The first SNN-oriented semi-structured pruning method.} We introduce \textbf{SpikeNM}, the first framework to learn \(N{:}M\) sparsity for SNNs \emph{from scratch}, enforcing \emph{at-most-\(N\)} non-zeros per \(M\)-weight block.
\item \textbf{Mask Search Complexity Linearization.} We replace subset learning over \(\sum_{k=1}^{N}\binom{M}{k}\) masks with an \(M\)-way \emph{basis-logit} parameterization and a differentiable top-\(k\) sampler, reducing per-block complexity to \(\mathcal O(M)\) and enabling more aggressive sparsification.
\item \textbf{Eligibility-inspired distillation regularizer.} We propose \emph{EID} to optimize the basis-logit probabilities via temporally accumulated credits, which improves mask-dynamics alignment. Extensive experiments show that under \(2{:}4\) sparsity most datasets exhibit no accuracy loss and sometimes even gains, as demonstrated in Figure~\ref{main}.
\end{itemize}

\section{Related Work}

\noindent\textbf{Spiking Neural Networks (SNNs)} process information with discrete spikes and temporal codes, delivering energy and latency benefits on neuromorphic hardware \citep{merolla2014million, Davies2018loihi, akopyan2015truenorth}. 
High performance is pursued via either \emph{ANN$\rightarrow$SNN conversion}, calibrating activations and simulation steps to reduce conversion error and time steps \citep{rueckauer2017conversion,deng2021optimal,bu2022optimalconv}, or \emph{direct training} with surrogate gradients\citep{neftci2019surrogate}, where STBP \citep{Wu2018STBP} enables deep optimization.
Recent work scales the depth of SNNs for performance improvement,  \emph{SEW-ResNet} \citep{fang2021deep} introduces spike-element-wise residual coupling to stabilize deep residual training and enable much deeper spiking backbones, \emph{Spikformer} \citep{zhou2022spikformer} adapts Transformer blocks to SNNs with spiking patch splitting and spiking self-attention for long-range dependency modeling
\emph{Spike-Driven Transformer v3} \citep{yao2025scaling} streamlines spiking attention with latency-aware modules and improved training stability for very deep stacks. However, these deeper architectures substantially increase parameter footprints and energy consumption, which is unfavorable for edge deployment, which makes network pruning necessary.

\vspace{1mm}
\noindent\textbf{Pruning for SNNs.}
Existing SNN pruning comprises two families. \emph{Structured} methods remove channels, filters, or neurons to form hardware-friendly dense tensors and simplify deployment. \emph{SCA} \citep{li2024towards} prunes channels by measured spiking activity, \emph{Network Slimming} \citep{li2024efficient} prunes channels using learned scaling factors (e.g., BN \(\gamma\)) and spiking activity to remove weak channels, followed by fine-tuning to recover accuracy.
These methods are constrained to the channel level, so they cannot exploit fine-grained within-kernel redundancy and typically achieve lower sparsity at a given accuracy. 
Moreover, their activity, BN-based saliency proxies are sensitive to input- and time-dependent spiking, risking the removal of informative low-rate channels and necessitating heavier fine-tuning.
\emph{Unstructured} methods instead zero individual weights by magnitude or saliency. \emph{Gradient Rewiring}~\citep{GradR} alternates prune-regrow steps using gradient-guided saliency to learn sparse connectivity; \emph{Dynamic Spatio-Temporal Pruning} \citep{gou2025dynamic} updates masks over time from spike/temporal redundancy to remove superfluous weights; \emph{UPR} \citep{UPR} targets energy by pruning weights and neurons under SynOps objectives. 
Despite high sparsity with minimal architectural change, their \emph{irregular patterns} hinder acceleration on general hardware and often depend on bespoke sparse kernels.
To reconcile these trade-offs, Semi-structured pruning was proposed for ANNs.

\vspace{1mm}
\noindent\textbf{Semi-Structured $N{:}M$ Sparsity.}
\noindent
With the advent of Sparse Tensor Cores in the NVIDIA Ampere architecture~\citep{mishra2021accelerating}, research on learning N{:}M semi-structured sparsity masks from scratch has flourished. A representative starting point, \emph{Learning N{:}M Fine-grained Structured Sparse Neural Networks From Scratch}~\citep{zhou2021learning}, refines the straight-through estimator under N{:}M constraints, and subsequent works explore adaptive N{:}M ratios across layers and training steps~\citep{sun2021dominosearch, lu2023step}. However, these promising directions remain largely unexplored for spiking neural networks (SNNs). Moreover, in pattern-as-class formulations the mask search space grows combinatorially with $M$ (i.e., $K=\binom{M}{N}$), which hinders deploying the higher sparsity regimes that SNNs could otherwise tolerate. Motivated by this gap, we investigate bringing N{:}M pruning to SNNs and propose strategies that reduce the mask exploration complexity while preserving the hardware-friendly benefits of semi-structured sparsity.

\section{Preliminary}
\label{sec:pre}

\noindent\textbf{Spiking Neuron Model} replace continuous activation function (\textit{e.g.,} ReLU) with spike neurons, enabling spike-driven sparsity and temporal state via membrane dynamics and resets. 
In this work, we adopt the discrete-time Leaky Integrate-and-Fire (LIF) neuron~\citep{LIF}.
The membrane potential and spike firing of the LIF model are governed by
\begin{align}
\tilde u_{t+1}^i
&= \alpha\, u_t^i \;+\; \sum\nolimits_j W^{ij} o_t^j,
\label{eq:lif-pre} \\
o_{t+1}^i
&= H(\tilde u_{t+1}^i - V_{\mathrm{th}}),
\label{eq:lif-spike} \\
u_{t+1}^i
&= \tilde u_{t+1}^i - V_{\mathrm{th}}\, o_t^i ,
\label{eq:lif-post}
\end{align}
where $\alpha = 1-\tfrac{1}{\tau}$ is the leaky factor, $H(\cdot)$ is the Heaviside function, $V_{\mathrm{th}}$ is the threshold, $W^{ij}$ are synaptic weights, and $u_t^i$ and $o_t^i$ denote the membrane potential and binary spike output of neuron $i$ at time step $t$, respectively.

\vspace{1mm}
\noindent\textbf{Spatio-Temporal Backpropagation (STBP) of SNNs.}
For SNNs, a major challenge is the non-differentiability introduced by spike generation. As shown in Eq.~\ref{eq:lif-spike}, spikes are produced by the Heaviside function, so gradients vanish almost everywhere except at threshold-crossing instants. To address this issue, direct SNN training typically adopts STBP~\citep{Wu2018STBP}. The surrogate function $\phi$ (\textit{e.g.,} sigmoid) enables gradient flow by approximating the spike derivative:
\begin{align}
\tfrac{\partial o_t^i}{\partial \tilde u_{t+1}^i}
\;\approx\;
\phi'\!\big(\tilde u_{t+1}^i - V_{\mathrm{th}}\big).
\end{align}
Let $\mathcal L = \sum_{t=1}^{T} \ell(\mathbf o_t, y)$ denote the task loss over all time steps.
STBP defines the spike and membrane error signals:
\begin{align}
e_t^i &= \tfrac{\partial \mathcal L}{\partial o_t^i}, \quad
\delta_t^i= \tfrac{\partial \mathcal L}{\partial \tilde u_{t+1}^i}.
\end{align}
The membrane error $\delta_t^i$ updates recursively per time step:
\begin{align}
\label{delta}
\!\!\!\delta_t^i
\!=\!\!
\Big(
   e_t^i
   \!+\!
   \sum\nolimits_k \!W^{i k}\delta_t^{k}
\Big)
\phi'\!\big(\tilde u_{t+1}^i \!-\! V_{\mathrm{th}}\big)
\!+\! \alpha\big(1 \!-\! o_t^i\big)\delta_{t+1}^i .
\end{align}
The final gradient w.r.t. synaptic weights is accumulated by:
\begin{align}
\frac{\partial \mathcal L}{\partial W^{ij}}
= \sum_{t=1}^{T} \delta_t^i\, o_t^j.
\label{stbp}
\end{align}

\section{Methodology}
\label{sec:method}
We motivate and present \textbf{SpikeNM}, a learnable semi-structured pruning method, which enforces $N{:}M$ sparsity in SNNs by jointly optimizing weights and probabilistic-based mask basis from scratch.
SpikeNM remains at most $N$ non-zero entries within each block of $M$ consecutive weights. 
\subsection{N:M Sparsity and Key Challenges}
The fundamental idea of N:M sparsity lies in dividing the entire network into non-overlapping blocks of $M$ consecutive parameters and then retaining at most $N$ effective weights for each block. Formally, this task can be characterized as a mask selection problem with the candidate set $\mathcal{S}^{N{:}M}\triangleq\{\mathcal{M}\in\{0,1\}^M: 1\le\|\mathcal M_i\|_0\le N\}$. For instance, when considering $B$ disjoint blocks $\{\mathcal W_i\in\mathbb R^{1\times M}\}_{i=1}^B$, the goal of N:M sparsity is to determine the optimal binary masks $\mathcal{M}^{*}_i$ from $\mathcal{S}^{N{:}M}$ such that the pruned weights $\{\mathcal{M}^{*}_i\odot\mathcal{W}_i \}_{i=1}^B$ can maintains its generalization capability on observed data, where $\odot$ denotes the Hadamard product. Thus, it is natural to define the following combinatorial optimization problem for N:M sparsity:
\begin{equation}\label{eq:prob}
\{\mathcal{M}_{i}^{*}\}=\mathop{\arg\min}_{\{\mathcal M_i\in\mathcal{S}^{N{:}M}\}}\mathbb E_{\substack{x\sim p(x)}}
\Big[
\mathcal L\big(x;\,\mathcal W_i \odot \mathcal M_i\big)
\Big],
\end{equation} where $p(x)$ denotes the underlying data distribution.

\vspace{1mm}
\noindent\textbf{Key Challenges for Eq.~\ref{eq:prob}.} Firstly, in order to effectively solve the Eq.~\ref{eq:prob}, we have to face a huge search space with exponential size of $|\mathcal{S}^{N{:}M}|$. Here, $|\mathcal{S}^{N{:}M}|$ denotes the cardinality of the mask set $\mathcal{S}^{N{:}M}$. For 2:4 sparsity, $|\mathcal S^{2{:}4}|=\sum_{k=1}^{2}{4\choose k}=10$, namely:
\begin{equation}
\begin{split}
\mathcal S^{2{:}4}\!\triangleq\!\{\!&
\texttt{[\!0 \!0 \!0\! 1\!]},\texttt{[\!0 \!0 \!1\! 0\!]},\ \ldots, \texttt{[\!1 \!0 \!0 \!0\!]},\\
&\texttt{[\!1 \!1 \!0 \!0\!]},\texttt{[\!1 \!0 \!1 \!0\!]},\ \ldots,\  \texttt{[\!0 \!0 \!1 \!1\!]}\!\}.
\end{split}
\end{equation}
This number grows rapidly with $M$, \textit{e.g.,} for a $2{:}8$ sparsity the space reaches $\sum_{k=1}^{2}\binom{8}{k}=36$, making optimizing Eq.~\ref{eq:prob} difficult and time-consuming. Moreover, the discrete nature of Eq.~\ref{eq:prob} limits the applicability of the well-established gradient-based methods such as SGD~\citep{boyd2004convex,lan2020first} to find the optimal masks $\{\mathcal{M}_{i}^{*}\}$. To overcome these issues, we introduce a novel probabilistic framework for Eq.~\ref{eq:prob}.

\subsection{Probabilistic N:M Mask}
\label{sec:mask-space}
In this subsection, we present the details of our proposed probabilistic N:M framework. Before that, we need to define a new operation $\oplus$ for the probabilistic sum~\citep{reza1994introduction,buchbinder2024constrained} of two vectors $\mathbf{a}$ and $\mathbf{b}$, namely, $\mathbf{a}\oplus\mathbf{b}=\mathbf{1}-(\mathbf{1}-\mathbf{a})\odot(\mathbf{1}-\mathbf{b})$ where $\odot$ is coordinate-wise product and $\one$ denotes $M$-dimensional vector whose all coordinates are $1$. Observe that $\oplus$ is a symmetric associative operator, and therefore, it makes sense to apply it also to sets of vectors. Formally, given multiple vectors $\mathbf{a}_{1},\dots,\mathbf{a}_{N}$, we also can define:
\begin{equation}
\bigoplus_{i=1}^{N}\mathbf{a}_{i}\triangleq\mathbf{a}_{1}\oplus\mathbf{a}_{2}\oplus\dots\oplus\mathbf{a}_{N}=\Bigg(\mathbf{1}-\bigodot_{i=1}^{N}\Big(1-\mathbf{a}_{i}\Big)\Bigg).
\label{eq:bigoplus}
\end{equation}
With this operation $\oplus$, we can get the following representation theorem for the N:M mask space $\mathcal{S}^{N:M}$, namely,
\begin{tcolorbox}[colback=SeaGreen!10!CornflowerBlue!10,colframe=RoyalPurple!55!Aquamarine!100!]
\begin{theorem}[Representation of $N{:}M$ sparsity]
\label{thm:rep}
\begin{equation*}
\label{eq:rep}
\mathcal S^{N{:}M}\triangleq\Big\{\;\bigoplus_{k=1}^{N}\mathbf a_k\;:\;\mathbf a_k\in\{\mathbf e_1,\dots,\mathbf e_M\}\;\Big\}.
\end{equation*}
where $\mathbf{e}_{j}$ is $j$-th basis vector in $\mathbb{R}^{M}$. $\bigoplus_{k=1}^{N}\mathbf a_k$ has exactly $N$ ones if $\mathbf a_k$ chooses distinct basis.
%{If the draws are with replacement and $\mathbf 0$ is allowed, then }$\big\|\bigoplus_{k=1}^{N}\mathbf a_k\big\|_0 \le N$ {(i.e., at most $N$ ones).}
\end{theorem}
\end{tcolorbox}
\noindent
Motivated by Theorem~\ref{eq:rep}, if we represent each mask $\mathcal{M}_{i}\in\mathcal{S}^{N:M}$ as a probabilistic sum of $\{\mathbf{a}_{i,1},\dots,\mathbf{a}_{i,N}\}$ where $\mathbf{a}_{i,j}\in\{\mathbf{e}_{1},\dots,\mathbf{e}_{M}\},\forall j\in[N]$, then we naturally can reformulate the mask selection Eq.~\ref{eq:prob} as follows:
\begin{equation}\label{eq:prob1}
\min_{\mathbf{a}_{i,j}\in\{\mathbf{e}_{1},\dots,\mathbf{e}_{M}\}}\mathbb E_{\substack{x\sim p(x)}}
\Big[
\mathcal L\big(x;\,\bigoplus_{j=1}^{N}\mathbf{a}_{i,j}\odot\mathcal W_i\big)
\Big].
\end{equation}
Eq.~\ref{eq:prob1} provides a compressed characterization for the Eq.~\ref{eq:prob}. Notably, encoding $N$ distinct vectors $\{\mathbf{a}_{i,1},\dots,\mathbf{a}_{i,N}\}$
generally involves up to $N\cdot M$ unknown parameters.  By comparison, the combinatorial mask set $\mathcal{S}^{N:M}$ often has a enormous size of $\big|\mathcal{S}^{N{:}M}\big|\triangleq\sum_{k=1}^{N}\binom{M}{k}$. Particularly when $N$ approaches $M$, the linear parameter growth $N\cdot M$ of vectors $\{\mathbf{a}_{i,1},\dots,\mathbf{a}_{i,N}\}$ can be significantly smaller than the exponential complexity of $\mathcal{S}^{N:M}$.
\vspace{1mm}
\noindent\textbf{Probabilistic Parameterization.}
It is worth noting that Eq.~\ref{eq:prob1} still is a discrete problem such that we cannot leverage gradient-based methods to search for the optimal masks.  To resolve this limitation, we next propose a novel probabilistic parameterization for Eq.~\ref{eq:prob1}.  Notably, the goal of  Eq.~\ref{eq:prob1} is to identify the optimal vector $\mathbf{a}_{i,j}$ from the basis  candidate set $\{\mathbf{e}_{1},\dots,\mathbf{e}_{M}\}$, which inherently resembles a sampling process over $\{\mathbf{e}_{1},\dots,\mathbf{e}_{M}\}$. Building on this insight, we  assign \emph{one} categorical distribution $\tilde{\pi}_i\triangleq(\tilde{\pi}_{i,1},\dots,\tilde{\pi}_{i,M})$ for every block $i$, where each $\tilde{\pi}_{i,j}$ represents the probability of selecting basis $\mathbf{e}_{j}$ and $\sum_{j=1}^{M}\tilde{\pi}_{i,j}=1$. After that, we utilize every categorical distribution $\tilde{\pi}_i$ to independently generate $N$ random basis vectors $\{\tilde{\mathbf{e}}_{i,1},\dots,\tilde{\mathbf{e}}_{i,N}\}$ where $\tilde{\mathbf{e}}_{i,j}\in\{\mathbf{e}_{1},\dots,\mathbf{e}_{M}\}$ and allocate them to $\{\mathbf{a}_{i,1},\dots,\mathbf{a}_{i,N}\}$ by setting $\mathbf{a}_{i,k}\triangleq\tilde{\mathbf{e}}_{i,k}$. 

With the aforementioned  probabilistic parameterization, we can naturally transfer the discrete problem Eq.~\ref{eq:prob1} into a continuous optimization over $\tilde{\pi}_i$ as follows:
\begin{equation}
\label{eq:prob-relax}
\begin{aligned}
\min_{\{\tilde{\pi}_i:\,\sum_{s=1}^M \tilde{\pi}_{i,s}=1\}}& \mathbb E_{\substack{x\sim p(x)\\ \mathbf a_{i,k}\sim \tilde{\pi}_i}}
\Big[
\mathcal L\big(x;\bigoplus_{k=1}^{N}\mathbf{a}_{i,k}\odot\mathcal W_i\big)
\Big],
\end{aligned}
\end{equation} where $a_{i,k}\sim \tilde{\pi}_i$ indicates independent sampling from the categorical distribution $\tilde{\pi}_i$ over the basis set $\{\mathbf{e}_{1},\dots,\mathbf{e}_{M}\}$.

\vspace{1mm}
\noindent
\textbf{Why this parameterization?} (i) \emph{Parameter efficiency:} Compared to the combinatorial complexity $\binom{M}{N}$ of the state-of-the-art work~\cite{maskllm}, our Eq.~\ref{eq:prob-relax} only needs to learn $M$ probabilities for every block $i$. Thus, by introducing randomness, the parameter scale of Eq.~\ref{eq:prob-relax} can be further reduced from the previous $N\cdot M$ of Eq.~\ref{eq:prob1} to a linear $M$. (ii) \emph{Lossless Mask Recovery:} For any categorical distribution $\tilde{\pi}_i$, we can, through Eq.~\ref{eq:prob-relax}, easily produce a N:M mask without any loss in terms of the function value in expectation.

\vspace{1mm}
\noindent\noindent\textbf{Differentiable Approximation for Eq.~\ref{eq:prob-relax}.} From the definition of Eq.~\ref{eq:prob-relax}, we derive the following equality:
\begin{equation}
\label{eq:obj_eq}
\!\!\!\!\begin{aligned}
&\mathbb E_{\substack{x\sim p(x)\\ \mathbf a_{i,k}\sim \tilde{\pi}_i}}
\Big[
\mathcal L\big(x;\bigoplus_{k=1}^{N}\mathbf{a}_{i,k}\odot\mathcal W_i\big)
\Big]\\&\!\!=\!\!\!\sum_{\mathbf{a}_{i,k}\in\{\mathbf{e}_{1},\dots,\mathbf{e}_{M}\}}\!\!\!\!\!\!\mathbb E_{\substack{x}}
\!\Big[
\mathcal L\big(x;\bigoplus_{k=1}^{N}\mathbf{a}_{i,k}\odot\mathcal W_i\big)\!\Big]\!\prod_{i,j}p(\mathbf{a}_{i,j}|\tilde{\pi}_{i}),
\end{aligned}
\end{equation} where $p(\mathbf{e}_{k}|\tilde{\pi}_{i})=\tilde{\pi}_{i,k},\forall k\in\{1,\dots,M\}$. Notably, the objective in Eq.~\ref{eq:obj_eq} is an exponential sum of polynomial terms parameterized by  $\tilde{\pi}_{i}$, which makes the gradient computation of Eq.~\ref{eq:obj_eq}  intractable. To enable gradient-based optimization via automatic differentiation, we next propose a differentiable approximation for Eq.~\ref{eq:obj_eq}.

A standard approach to model a  categorical sampling is the gumbel-max~\citep{gumbel}, a re-parameterization trick that decouples the randomness of sampling into a noise variable. Specifically, for each block $i$, we can draw basis vector from the categorical $\tilde{\pi}_i$ by an auxiliary noise variable $\epsilon$, it produces the one-hot vector $\mathbf y_{i,k},\forall k\in\{1,\dots,N\}$ as follows:
\begin{equation}
\begin{aligned}
\mathbf y_{i,k}(\tilde{\pi}_{i}) &\triangleq \operatorname{onehot}\!\big(\arg\!\!\!\!\!\!\max_{s\in\{1,\dots,M\}}\![\log(\tilde{\pi}_{i,s})+g_{i,k,s}]\big),\\ g_{i,k,s} &\triangleq -\log(-\log \epsilon_{i,k,s}),\quad \epsilon_{i,k,s}\sim U(0,1),
\end{aligned}
\label{eq:gumbel_max}
\end{equation}
where $U(0,1)$ represent the uniform distribution over $(0,1)$ and $g_{i,s} $ is known as the Gumbel noise. With this Gumbel-Max trick, the original Eq.~\ref{eq:prob-relax} can can be reformulated as:
\begin{equation}
\label{eq:prob-relax1}
\begin{aligned}
\!\!\!\!\min_{\{\tilde{\pi}_i:\sum_{s} \!\!\tilde{\pi}_{i,s}=1\}}\!\!\mathbb E_{\!\!\substack{x\sim p(x)\\\epsilon_{i,k,s}\sim U(0,1)}}
\!\!\Big[
\mathcal L\!\big(x;\bigoplus_{k=1}^{N}\mathbf{y}_{i,k}(\tilde{\pi}_{i})\!\odot\!\mathcal W_i\big)
\!\Big].
\end{aligned}
\end{equation}
Eq.~\ref{eq:prob-relax1} shifts the optimization variables into the argument of the loss function $\mathcal L$, enabling gradient computation via automatic differentiation. However, the $\arg\max$ and one-hot
operations in gumbel-max trick are inherently non-differentiable. 
% To address this, we adopt the Straight-Through Gumbel-Softmax (STGS)~\cite{jang2017categorical} which adopts the one-hot vector $\mathbf y_{i,k}$ in Eq.~\ref{eq:gumbel_max} in the forward process and approximate the one-hot vector with Softmax, yielding a soft and differentiable     
% $\tilde{\mathbf{y}}_{i,k}=[\tilde{y}_{i,k,1},\ldots,\tilde y_{i,k,M}]$, defined in Eq.~\ref{eq:gumbel-softmax} below, in the backward process for gradient backpropagation
% \begin{equation}
% \label{eq:gumbel-softmax}
% \tilde{y}_{i,k,s}\triangleq\frac{\exp\big((\log(\tilde{\pi}_{i,s})+g_{i,k,s})/\tau\big)}
%         {\sum_{j}\exp\big((\log(\tilde{\pi}_{i,j})+g_{i,k,j})/\tau\big)},
% \end{equation}
To this end, we adopt the Straight-Through Gumbel--Softmax (STGS)~\cite{jang2017categorical}: 
the forward pass employs the one-hot sample $\mathbf{y}_{i,k}$ from Eq.~\ref{eq:gumbel_max}, 
while the backward pass replaces it with a Softmax relaxation,
yielding a differentiable vector 
$\tilde{\mathbf{y}}_{i,k}=[\tilde y_{i,k,1},\ldots,\tilde y_{i,k,M}]$ for gradient flow:
\begin{equation}
\label{eq:gumbel-softmax}
\tilde{y}_{i,k,s}
=\frac{\exp\!\big((\log \tilde{\pi}_{i,s}+g_{i,k,s})/\tau\big)}
        {\sum_{j}\exp\!\big((\log \tilde{\pi}_{i,j}+g_{i,k,j})/\tau\big)},
\end{equation}
where $\tau$ is the predefined temperature term. As a result, we get the following differentiable approximation for Eq.~\ref{eq:prob-relax}:
\begin{equation}
\label{eq:prob-relax2}
\begin{aligned}
&\!\!\!\!\min_{\{\tilde{\pi}_i:\sum_{s} \!\!\tilde{\pi}_{i,s}=1\}}\!\!\mathbb E_{\!\!\substack{x\sim p(x)\\\epsilon_{i,k,s}\sim U(0,1)}}
\!\!\Big[
\mathcal L\big(x;\bigoplus_{k=1}^{N}\hat{\mathbf{y}}_{i,k}(\tilde{\pi}_{i})\!\odot\!\mathcal W_i\big)
\Big]\\
&\quad\text{where}~~~\hat{\mathbf{y}}_{i,k}=\tilde{\mathbf{y}}_{i,k}-\mathrm{stopgrad}(\tilde{\mathbf{y}}_{i,k}-{\mathbf{y}}_{i,k}),
\end{aligned}
\end{equation} where $\mathrm{stopgrad}$ denotes stop gradient operator.

\vspace{1mm}
\noindent\textbf{Temperature annealing.}
In STGS, $\tau$ controls the entropy of the categorical relaxation: larger $\tau$ yields smoother, low-confidence soft indices, while smaller $\tau$ produces peaky, high-confidence ones. To transition from exploration to near-discrete selection, we anneal $\tau$ geometrically as:
\begin{equation}
\label{eq:tau}
\tau_t
=
\max\!\Big(
\tau_{\min},\ 
\tau_{\max}\!\cdot\!\big(\tfrac{\tau_{\min}}{\tau_{\max}}\big)^{\,t/T_{\text{search}}}
\Big),
\end{equation} where $t$ is the search step and $T_{\text{search}}$ the total search iteration for mask optimizing. As $\tau\!\downarrow\!0$, the relaxation sharpens toward one-hot picks, aligning the learned logits with a concrete N:M mask at the end of search.

\vspace{1mm}
\noindent\textbf{Unconstrained Reformulation.} Finally, we utilize a reformulation trick to eliminate the unit constraint in Eq.~\ref{eq:prob-relax2}, namely, $\sum_{s=1}^M \tilde{\pi}_{i,s}=1$. This step is crucial as it enables us to avoid the computationally expensive projection operations. Specifically, we reset $\tilde{\pi}_{i,s}\triangleq\text{softmax($\boldsymbol{\theta}_i$)}$ where $\boldsymbol{\theta}_i=(\theta_{i,1},\dots,\theta_{i,M})$ is the logits of softmax function. With this tool, we can transform the Eq.~\ref{eq:prob-relax2} as an unconstrained continuous optimization regarding the logits $\boldsymbol{\theta}_i$, that is, 
\begin{equation}
\label{eq:un-prob-relax}
\begin{aligned}
&\min_{\boldsymbol{\theta}_i}\mathbb{E}_{\substack{x\sim p(x)\\\epsilon_{i,k,s}\sim U(0,1)}}
\Big[
\mathcal L\big(x;\bigoplus_{k=1}^{N}\hat{\mathbf{y}}_{i,k}(\boldsymbol{\theta}_i)\odot\mathcal W_i\big)
\Big]\\
&\text{where}~~~\hat{\mathbf{y}}_{i,k}=\tilde{\mathbf{y}}_{i,k}-\mathrm{stopgrad}(\tilde{\mathbf{y}}_{i,k}-{\mathbf{y}}_{i,k}).
\end{aligned}
\end{equation}

% \subsection{Eligibility-Inspired Self Distillation}
% \label{sec:eid}
% Unlike ANNs, SNN activations are \emph{sparse} and \emph{time-varying}; synaptic importance is therefore highly nonstationary across timesteps. When learning an $N{:}M$ mask jointly with weights, this can cause a \emph{mask-weight mismatch}: the mask optimizer receives high-variance, delayed signals from the task loss, while the most causally useful synapses at a given time are those that contribute spike-wise and temporally. Inspired by eligibility traces, we optimize the mask via a self-distillation scheme guided by eligibility-based temporal credit.

% \vspace{1mm}
% \noindent\textbf{Distillation between eligibility-inspired credit and basis logits.}
% Inspired by eligibility traces, we use the STBP-consistent term $\delta_t^i o_t^j$ as a practical proxy for synapse-wise eligibility, measuring each basis’ temporal contribution. We aggregate per-block credit as
% \begin{equation}
% s_{i,m}=\sum_{t=1}^{T}\big|\delta_t^{\,i}o_t^{\,m}\big| .
% \end{equation}
% We then convert these credits into a probability (target) distribution via softmax:
% \begin{equation}
% q_{i,s}
% =\frac{\exp(s_{i,s}/\tau_q)}{\sum_{r=1}^{M}\exp(s_{i,r}/\tau_q)} ,
% \quad s=1,\dots,M .
% \end{equation}
% Let $p_{\pi_i}$ be the mask distribution obtained from the basis logits in block $i$. We perform a single KL-based self-distillation to align the two:
% \begin{equation}
% \mathcal L_{\mathrm{EID}}
% =\sum_{i}\mathrm{KL}\!\big(q_i\;\|\;p_{\pi_i}\big) .
% \end{equation}

\subsection{Eligibility-Inspired Distillation (EID)}
\label{sec:eid}
Unlike ANNs, SNN activations are \emph{sparse} and \emph{time-varying}, so synaptic importance is nonstationary across timesteps. When learning $N{:}M$ masks jointly with weights, direct task supervision is delayed and high-variance, causing a \emph{mask--weight mismatch}. We therefore distill a temporally grounded target into the per-block mask distribution to stabilize mask learning.

\vspace{1mm}\noindent\textbf{From neuroscience to a practical credit signal.}
An \emph{eligibility trace} is a \emph{local} synaptic signal quantifying how a weight modulates postsynaptic dynamics, later gated by a task-dependent learning signal~\citep{fremaux2016,bellec2020prop}. Under STBP (Sec.~3), the instantaneous per-weight gradient plays this role: for synapse $(j\!\!\to\!\! i)$ at time $t$, the gradient $g^{\,t}_{ij}$ w.r.t.\ $W^{ij}$ satisfies $g^{\,t}_{ij}\!=\!\delta_t^{\,i}o_t^{\,j}$ (cf.\ Eq.~\ref{stbp}). Eligibility and gradient magnitude are equivalent up to time-dependent scaling (proof in the appendix). We thus accumulate magnitudes over time $C_{ij}\!=\!\sum_{t=1}^{T}\!\big|g^{\,t}_{ij}\big|$ as a practical proxy for eligibility.

\begin{algorithm}[t]
\caption{SpikeNM: Learning $N{:}M$ Structured SNN}
\label{alg:SpikeNM}
\begin{algorithmic}[1]
\REQUIRE Dataset $\mathcal{D}$; blocks $B$; group size $M$; budget $N$; search epochs $T_{\text{search}}$; finetune epochs $T_{\text{ft}}$; temperatures $(\tau_{\max},\tau_{\min})$; EID params $(\lambda,\tau_{q})$.
\STATE \textbf{Init:} Randomly init weights $\{\mathcal{W}_i\}_{i=1}^{B}$ and per-block logits $\{\boldsymbol{\theta}_i\in\mathbb{R}^{M}\}_{i=1}^{B}$.\\
\vspace{0.5mm}
\textcolor{teal}{\# Searching stage (optimize weights and masks)}
\vspace{0.5mm}
\FOR{$t=1$ \TO $T_{\text{search}}$}
  \STATE $\displaystyle \tau \leftarrow \max\!\Big(\tau_{\min},\,\tau_{\max}\big(\tfrac{\tau_{\min}}{\tau_{\max}}\big)^{t/T_{\text{search}}}\Big)$ \hfill \textcolor{teal}{\# Eq.~\ref{eq:tau}}
  \FOR{mini-batch data $x_{\mathcal{B}} \subset \mathcal{D}$ of batch size $\mathcal{B}$}
    \STATE $\hat{\mathbf{y}}_{i,k}=\tilde{\mathbf{y}}_{i,k}-\mathrm{stopgrad}(\tilde{\mathbf{y}}_{i,k}-{\mathbf{y}}_{i,k})$ \\$\displaystyle \mathcal{L}_{\text{task}} \leftarrow \frac{1}{\mathcal{B}}\mathcal{L}\big(x_{\mathcal{B}};\bigoplus_{k=1}^{N}\hat{\mathbf{y}}_{i,k}(\boldsymbol{\theta}_i)\!\odot\!\mathcal W_i\big)$ \hfill \textcolor{teal}{\# Eq.~\ref{eq:un-prob-relax}}
    \STATE $\mathcal{L}_{\mathrm{EID}}=\sum_{i=1}^{B}\mathrm{KL}\big(q_i\,\|\,\tilde{\pi}_i(\boldsymbol{\theta}_i)\big)$ \hfill \textcolor{teal}{\# Eq.~\ref{eq:eid-kl}}
    \vspace{1mm}
    \STATE Minimize $\mathcal{L}_{\text{task}}\!+\!\lambda\mathcal{L}_{\mathrm{EID}}$ over $\{\mathcal{W}_i\},\{\boldsymbol{\theta}_i\}$  \textcolor{teal}{\# Eq.~\ref{eq:overall_optimization}}
  \ENDFOR
\ENDFOR\\
\vspace{0.5mm}
\textcolor{teal}{\# Pruning stage (instantiate a hard $N{:}M$ subnetwork))}
\vspace{0.5mm}
\STATE Set $\displaystyle \hat{\mathcal M}_i \leftarrow \bigoplus_{k=1}^{N}\mathbf{y}_{i,k}$,~$\mathcal W_i \leftarrow \hat{\mathcal M}_i \odot \mathcal W_i$;\ \textbf{freeze} $\{\hat{\mathcal M}_i\}$.\\
\vspace{1mm}
\textcolor{teal}{\# Finetuning stage (optimize subnetwork only))}
\vspace{1mm}
\STATE Minimize ${\mathbb{E}}_{x\sim p(x)}\big[\mathcal{L}(x;\,\hat{\mathcal M}_i \odot \mathcal W_i)\big]$ for $T_{\text{ft}}$ epochs w.r.t.\ $\{\mathcal W_i\}$.
\end{algorithmic}
\end{algorithm}

\vspace{1mm}
\noindent\textbf{From per-weight credits to $N{:}M$ block targets.}
We first compute $C_{ij}$ for all parameters $W^{ij}$ from all layers, then map each weight to its $N{:}M$ block (as in Sec.~4.1--4.2): if $W^{ij}$ belongs to block $i$ at within-block position $s\!\in\!\{1,\dots,M\}$, we set
$\mathcal{C}_{i,s}\;=\;C_{ij}$.

\vspace{1mm}
\noindent\textbf{Blockwise soft targets and KL alignment.}
Within each $N{:}M$ block, the aggregated credits $\mathcal{C}_{i}=\{\mathcal{C}_{i,s}\}_{s=1}^{M}$ encode \emph{relative} preference over positions. 
We convert them into a \emph{soft} target to preserve ranking while controlling confidence via a temperature $\tau_q$ :
\begin{equation}
\label{eq:eid-softmax}
q_{i,s}
\;=\;
\frac{\exp(\mathcal{C}_{i,s}/\tau_q)}
{\sum_{r=1}^{M}\exp(\mathcal{C}_{i,r}/\tau_q)} ,
\quad s=1,\dots,M .
\end{equation}
Let $\tilde{\pi}_i$ be the block-$i$ mask distribution induced by basis logits (Sec.~4.2). 
We align $\tilde{\pi}_i$ to $q_i$ using the KL loss, which penalizes \emph{under-allocation} to high-credit positions and reduces sampling variance as follows:
\begin{equation}
\label{eq:eid-kl}
\!\!\!\mathcal{L}_{\mathrm{EID}}
\!=\!
\frac{1}{B}\sum_{i=1}^{B}\mathrm{KL}\big(q_i\,\|\,\tilde{\pi}_i\big)
\!=\!
\frac{1}{B}\sum_{i=1}^{B}\sum_{s=1}^{M} q_{i,s}\,\log\!\frac{q_{i,s}}{\tilde{\pi}_{i,s}} .
\end{equation}

\subsection{Learning $N{:}M$ Structured SNN from Scratch}
\label{sec:from-scratch}
As shown in Alg.~\ref{alg:SpikeNM}, SpikeNM has 3 phases: \emph{search}, \emph{prune}, and \emph{finetune}. The search phase jointly learns task weights and per-block mask logits under a differentiable relaxation. The prune phase materializes a single hard $N{:}M$ mask per block from the last one in the search phase. The finetune phase then consolidates accuracy with masks frozen.

\vspace{1mm}
\noindent\textbf{Search (joint weights \& masks).}
We jointly optimize SNN weights $\{\mathcal{W}_i\}$ and per-block basis logits $\{\boldsymbol{\theta}_i\}$ using the unconstrained formulation in Eq.~\ref{eq:un-prob-relax} together with EID regularizer in Eq.~\ref{eq:eid-kl}. The overall objective is
\begin{equation}
\label{eq:overall_optimization}
\min_{\{\mathcal{W}_i\},\,\{\boldsymbol{\theta}_i\}}
\ \mathbb{E}_{\substack{x\sim p(x)\\ \epsilon_{i,k,s}\sim U(0,1)}}
\!\Big[
\mathcal L\!\big(x;\bigoplus_{k=1}^{N}\hat{\mathbf{y}}_{i,k}(\boldsymbol{\theta}_i)\odot\mathcal W_i\big)
\Big]
+\lambda\,\mathcal{L}_{\mathrm{EID}} .
\end{equation}
Across $T_{\text{search}}$ epochs, we anneal $\tau_t$ by Eq.~\ref{eq:tau}. This phase learns task-useful weights and confident mask per block.

\vspace{1mm}
\noindent\textbf{Prune (instantiate a hard $N{:}M$ subnetwork).}
At the end of search we do not re-sample. For each block $i$, we take the last hard one-hots $\{\mathbf{y}_{i,k}\}_{k=1}^{N}$ from the search phase and OR-compose them to obtain the mask
$\hat{\mathcal M}_i \;=\; \bigoplus_{k=1}^{N}\mathbf{y}_{i,k},~
\mathcal W_i \leftarrow \hat{\mathcal M}_i \odot \mathcal W_i$,
which materializes an at-most-$N{:}M$ subnetwork. We then freeze the masks $\{\hat{\mathcal M}_i\}$.

\vspace{1mm}
\noindent\textbf{Finetune (frozen masks, weights only).}
With masks fixed, we optimize the surviving weights using the task loss
\begin{equation}
\min_{\{\mathcal W_i\}}\ \mathbb E_{x\sim p(x)}\big[\mathcal L\big(x;\hat{\mathcal M}_i \odot \mathcal W_i\big)\big].
\end{equation}
This stabilizes training in the target $N{:}M$-sparse topology.

\begin{table*}[!t]
    \caption{Performance comparison between baselines. “--” denotes not reported.
    Acc denotes top-1 accuracy.
    \emph{$\triangle$ indicates the accuracy difference relative to the unpruned models. \(T\) is the number of time steps.}}
  \renewcommand{\arraystretch}{0.9}
  \vspace{-3mm}\label{tab:main_results}
  \centering
  \small
  \setlength{\tabcolsep}{3.5pt}
  \renewcommand{\arraystretch}{1.1}
  \begin{tabular}{@{} l l l l
  cccccc}
    \toprule
    \textbf{Dataset} & \textbf{Category} & \textbf{Method} &
    \textbf{Model} &
    \textbf{T} &
    \textbf{Acc} (\%) &
    {$\triangle$} ({\%}) &
    \textbf{SOPs} ({M}) &
    \textbf{Weight.} ({\%}) &
    {\textbf{Conn.} (\%)}\\
    \midrule

    % ================= CIFAR10 =================
    \multirow{7}{*}{\textbf{CIFAR10}}
      & Structured   & NSL \citep{li2024efficient} & VGGSNN \citep{VGGSNN}
        & \num{4}  & \num{91.22} & \num{-0.49} & \multicolumn{1}{c}{--} & \multicolumn{1}{c}{--} & \num{57.88} \\
      & Structured   & SCA-based \citep{li2024towards} & VGGSNN \citep{VGGSNN}
        & \num{4}  & \num{91.14} & \num{+0.32} & \multicolumn{1}{c}{--} & \multicolumn{1}{c}{--} & \num{55.86} \\
      & Unstructured & Grad R \citep{GradR} & 6Conv+2FC
        & \num{8}  & \num{92.54} & \num{-0.30} & \num{371.1} & \multicolumn{1}{c}{--}  & \num{36.72} \\
      & Unstructured & ESLSNN \citep{ESLSNN} & ResNet19 \citep{ResNet}
        & \num{2} & \num{91.09} & \num{-1.70} & \multicolumn{1}{c}{--} & \multicolumn{1}{c}{--} & \num{50.00} \\
      & Unstructured & UPR \citep{UPR}& 6Conv+2FC
        & \num{8}  & \num{92.63} & \num{-0.21} & {\num{38.3}} & {\num{34.24}} & {\num{2.97}} \\
      \cmidrule{2-10}
      & \textbf{Semi-Structured}   & \textbf{SpikeNM (2:4)} & {6Conv+2FC}
        & {\num{8}}  & {\bfseries \num{93.64}} & {\bfseries \num{+0.80}} & {\num{348.3}} & { \num{43.27}} & {\num{43.28}} \\
      & \textbf{Semi-Structured}   & \textbf{SpikeNM (2:8)} & {6Conv+2FC}
        & {\num{8}}  & {\bfseries \num{92.92}} & {\bfseries \num{+0.08}} & {\num{196.3}} & {\num{23.19}} & { \num{23.21}} \\
    \midrule

    % ================= CIFAR100 =================
    \multirow{6}{*}{\textbf{CIFAR100}}
      & Structured   & NSL \citep{li2024efficient} & ResNet18 \citep{fang2021deep}
        & \num{4}  & \num{66.40} & \num{-0.29} & \multicolumn{1}{c}{--} & \multicolumn{1}{c}{--} & \num{56.53} \\
      & Structured   & SCA-based \citep{li2024towards} & VGGSNN \citep{VGGSNN}
        & \num{4} & \num{64.89} & \num{+0.64} & \multicolumn{1}{c}{--} & \multicolumn{1}{c}{--} & \num{23.52} \\
      & Unstructured & ESLSNN \citep{ESLSNN}& ResNet19 \citep{ResNet}
        & \num{2} & \num{73.48} & \num{-0.99} & \multicolumn{1}{c}{--} & \multicolumn{1}{c}{--} & \num{50.00} \\
      & Unstructured & UPR \citep{UPR}& SEW-R18 \citep{fang2021deep}
        & \num{4}  & \num{72.34} & \num{-1.82} & {\num{27.2}} & {\num{29.35}} & {\num{11.76}} \\
      \cmidrule{2-10}
      & \textbf{Semi-Structured}   & \textbf{SpikeNM (2:4)} & {SEW-R18 \citep{fang2021deep}}
        & { \num{4}}  & {\bfseries \num{73.92}} & {\bfseries \num{-0.24}} & {\num{350.2}} & {\num{43.27}} & {\num{43.29}} \\
      & \textbf{Semi-Structured}   & \textbf{SpikeNM (2:8)} & {SEW-R18 \citep{fang2021deep}}
        & {\num{4}}  & {\bfseries \num{73.35}} & {\bfseries \num{-0.81}} & {\num{201.6}} & {\num{23.28}} & {\num{23.31}} \\
    \midrule

    % ================= CIFAR10-DVS =================
    \multirow{6}{*}{\textbf{CIFAR10-DVS}}
      & Structured   & SCA-based \cite{li2024towards} & 5Conv+1FC
        & \multicolumn{1}{c}{--} & \num{72.80} & \num{+0.90} & \multicolumn{1}{c}{--} & \multicolumn{1}{c}{--} & \num{21.73} \\
      & Unstructured & ESLSNN \citep{ESLSNN}& VGGSNN \citep{VGGSNN}
        & \multicolumn{1}{c}{--} & \num{78.30} & \num{-0.28} & \multicolumn{1}{c}{--} & \multicolumn{1}{c}{--} & \num{10.00} \\
      & Unstructured & STDS \cite{stds}& VGGSNN \citep{VGGSNN}
        & \num{10} & \num{81.70} & \num{-0.70} & \num{225.3} & \multicolumn{1}{c}{--} & \num{21.36} \\
      & Unstructured & UPR \citep{UPR}& VGGSNN \citep{VGGSNN}
        & \num{10} & \num{81.90} & \num{-0.50} & {\num{47.8}} & {\num{36.35}} & {\num{6.80}} \\
      \cmidrule{2-10}
      & \textbf{Semi-Structured}   & \textbf{SpikeNM (2:4)} & {VGGSNN} \citep{VGGSNN}
        & { \num{10}}  & {\bfseries \num{83.90}} & {\bfseries \num{+1.50}} & {\num{289.3}} & {\num{43.29}} & {\num{43.32}} \\
      & \textbf{Semi-Structured}   & \textbf{SpikeNM (2:8)} & {VGGSNN} \citep{VGGSNN}
        & { \num{10}}  & {\bfseries \num{83.20}} & {\bfseries \num{+0.80}} & {\num{165.3}} & {\num{25.58}} & {\num{24.70}} \\
    \midrule

    % ================= DVS-Gestures =================
    \multirow{2}{*}{\textbf{DVS-Gestures}}
      & Structured   & NSL \citep{li2024efficient} & VGG13 \citep{VGGSNN}
        & \num{10}  & \num{94.79} & \num{-0.61} & \multicolumn{1}{c}{--} & \num{72.90} & \multicolumn{1}{c}{--} \\
      \cmidrule{2-10}
      & \textbf{Semi-Structured}   & \textbf{SpikeNM (2:4)} & {VGG13 \citep{VGGSNN}}
        & { \num{10}}  & {\bfseries \num{95.73}} & {\bfseries \num{+0.33}} & {\bfseries \num{166.4}} & { \num{43.29}} & { \num{43.29}} \\
    \bottomrule
  \end{tabular}
  \vspace{-2mm}
\end{table*}

\section{Experiments}
\label{sec:exp_results}
We evaluate \textbf{SpikeNM} on RGB datasets (CIFAR10, CIFAR100~\citep{cifar10}) and neuromorphic datasets (CIFAR10-DVS~\citep{CIFAR10DVS}, DVS-Gestures~\citep{dvsgesture}) to validate its effectiveness. For CIFAR10, CIFAR100, and CIFAR10-DVS, we follow the UPR \citep{UPR} training and pruning configurations. For DVS-Gestures, which UPR does not report, we adopt the same configuration as CIFAR10-DVS. Experiments use PyTorch~\citep{pytorch} with SpikingJelly~\citep{spikingjelly} on 4 A40 GPUs. More details are in Sec.~\ref{exp_set} of the appendix.

\vspace{1mm}
\noindent\textbf{Metrics.} We take synaptic operations (SOPs) as the efficiency metric, with details in Sec.~\ref{sops} of the appendix.
In parallel, pruning strength is characterized by Weight. and Conn., which represent the proportions of weights and connectivity \emph{preserved} after pruning.

\subsection{Comparison to the State-of-the-Art}
\label{sec:perf-compare}
In this section, we compare \textbf{SpikeNM} against state-of-the-art SNN pruning methods, including structured approaches (NSL~\citep{li2024efficient}, SCA-based~\citep{li2024towards}) and unstructured approaches (Grad Rewiring~\citep{GradR}, ESLSNN~\citep{ESLSNN}, STDS~\citep{stds}, UPR~\citep{UPR}), with results summarized in Table~\ref{tab:main_results}. In particular, Table~\ref{tab:main_results} reports {SpikeNM} under $N\!:\!M$ of $2\!:\!4$ and $2\!:\!8$.

Across datasets and sparsity settings, SpikeNM delivers state-of-the-art accuracy with small (often positive) accuracy change while enforcing hardware-friendly \(N{:}M\) patterns on all the datasets. 
On \textbf{CIFAR10}, our \(2{:}4\) and \(2{:}8\) pruned models retain \(43\%\) and \(23\%\) of weights, yet reach \textbf{93.64\%} (\(+0.80\%\)) and \textbf{92.92\%} (\(+0.08\%\)), slightly surpassing the unpruned model and indicating nontrivial redundancy in the original SNN architecture.
On \textbf{CIFAR100}, at comparable sparsity, we obtain \textbf{73.92\%} (\(-0.24\%\)) and \textbf{73.35\%} (\(-0.81\%\)). Despite the modest accuracy change, our method still outperforms most of the baselines.
Neuromorphic datasets show the same trend. 
On \textbf{CIFAR10-DVS}, 
\textsc{SpikeNM} reaches \textbf{83.90\%} (\(+1.50\%\)) and \textbf{83.20\%} (\(+0.80\%\)), showing clear improvements over the unpruned model and baselines. This supports the view that SNNs admit stronger compressibility on sparse event-based inputs than on dense RGB data. 
On \textbf{DVS-Gestures}, the \(2{:}4\) configuration attains \textbf{95.73\%} (\(+0.33\%\)) compared with NSL (\(94.79\%\)), suggesting that tasks with richer temporal dynamics are more challenging to prune.

% \begin{figure}[!t]
%   \centering
%   \begin{subfigure}[t]{0.48\textwidth} 
%     \centering
% \includegraphics[scale=0.11]{vis_ablaEP.png}\caption{}\label{fig:ablaEP}
%   \end{subfigure}\\
%   \begin{subfigure}[t]{0.48\textwidth}
%     \centering
%     \includegraphics[scale=0.11]{vis_ablaEID.png}\caption{}\label{fig:ablaEID}
%   \end{subfigure}
%   \vspace{-4mm}
%   \caption{Ablation results on CIFAR10-DVS. (a) Ablation on different search epochs with a fixed total budget of 320 epochs (\(\text{search}+\text{finetune}=320\)). The dashed line marks the unpruned baseline (82.4\%). (b) Ablation on \(\lambda_{\mathrm{EID}}\). Bars show accuracies for non-zero \(\lambda\). The dashed line is \(\lambda=0\), denotes that no EID used.}
%   \label{fig:abla_combo}
%   \vspace{-4mm}
% \end{figure}

% 需要：
% \usepackage{array}
% \usepackage{subcaption}

In terms of \textbf{efficiency},
the global sparsity induced by the \(N{:}M\) scheme is approximately consistent across datasets: \(2{:}4\) retains \(\sim\!43\%\) connectivity, whereas \(2{:}8\) retains \(\sim\!23\%\). 
For SOPs related to energy consumption, \textsc{SpikeNM} remains comparable to the baselines (\textit{e.g.,} on \textbf{CIFAR10}: \(348.3\,\mathrm{M}\) vs.\ \(371.1\,\mathrm{M}\) for Grad~R), while preserving dense tensor layouts that are amenable to hardware acceleration.
Notably, \emph{UPR} reports the \emph{lowest} SOPs and Connectivity because it prunes \emph{neurons} in addition to weights, directly shrinking activation graphs and thus enjoying an inherent advantage on energy-style metrics. To separate the effect of neuron removal, we include a more detailed analysis and ablations under same neuron-pruning regimes in Table~\ref{tab:granu}.

\subsection{Ablation Studies}
\noindent\textbf{Ablation on Temperature annealing.}
Table~\ref{tab:temp_anneal} ablates the Gumbel-Softmax temperature by varying the start \(\tau_{\max}\) and end \(\tau_{\min}\). 
Moving from high to low \(\tau\) shifts sampling from smooth exploration to high-confidence choices. A fixed temperature (\(\tau_{\max}{=}\tau_{\min}{=}1\)) gives 82.4\%, while \emph{moderate} annealing performs best: \(\tau_{\max}{=}1,\ \tau_{\min}{=}10^{-1}\) reaches \textbf{83.9\%}. Colder ends help but are less stable (\textit{e.g.,} \(\tau_{\min}{=}10^{-3}\): 82.6\%; \(\tau_{\max}{=}10^{-1},\ \tau_{\min}{=}10^{-4}\): 83.2\%). Starting too cold harms exploration and accuracy (81.7-81.8\%). Empirically, runs that start warm (\(\tau_{\max}\!\approx\!1\)) and anneal to a modestly cold \(\tau_{\min}\!\in\![10^{-1},10^{-3}]\) achieve the strongest and most stable results. In this work, we set \(\tau_{\max}=1\) and \(\tau_{\min}=10^{-1}\) across all datasets.

\begin{table}[!t]
\centering
\caption{Ablation on Gumbel--Softmax temperature schedule: start ($\tau_{\max}$) vs end ($\tau_{\min}$) with CIFAR10-DVS and VGGSNN.}
\renewcommand{\arraystretch}{0.8}
\vspace{-3mm}
\begin{tabular}{lccc}
\toprule
Config & $\tau_{\max}$ (start) & $\tau_{\min}$ (end) & Acc (\%) \\
\midrule
A & $1$     & $1$ & 82.4 \\
B & $\textbf{1}$     & {\bfseries\boldmath $10^{-1}$} & \textbf{83.9} \\
C & $1$     & $10^{-2}$ & 83.5 \\
D & $1$     & $10^{-3}$ & 82.6 \\
E & $10^{-1}$ & $10^{-1}$ & 81.7 \\
F & $10^{-1}$ & $10^{-2}$ & 82.1 \\
G & $10^{-1}$ & $10^{-3}$ & 82.9 \\
H & $10^{-1}$ & $10^{-4}$ & 83.2 \\
I & $10^{-2}$ & $10^{-2}$ & 81.7 \\
J & $10^{-2}$ & $10^{-3}$ & 81.4 \\
K & $10^{-2}$ & $10^{-4}$ & 81.7 \\
L & $10^{-3}$ & $10^{-3}$ & 81.8 \\
M & $10^{-3}$ & $10^{-4}$ & 82.0 \\
N & $10^{-4}$ & $10^{-4}$ & 82.2 \\

\bottomrule
\end{tabular}
\label{tab:temp_anneal}
\vspace{-4mm}
\end{table}

\begin{figure}[!t]
  \centering
  % ----- (a) -----
  \hspace{-37mm}\begin{subfigure}[t]{0.48\textwidth}
    \centering
    \begin{tabular}{@{}m{0.01\textwidth} m{0.45\textwidth}@{}}
      \centering\hspace*{-0.5mm}(a) & \includegraphics[scale=0.125]{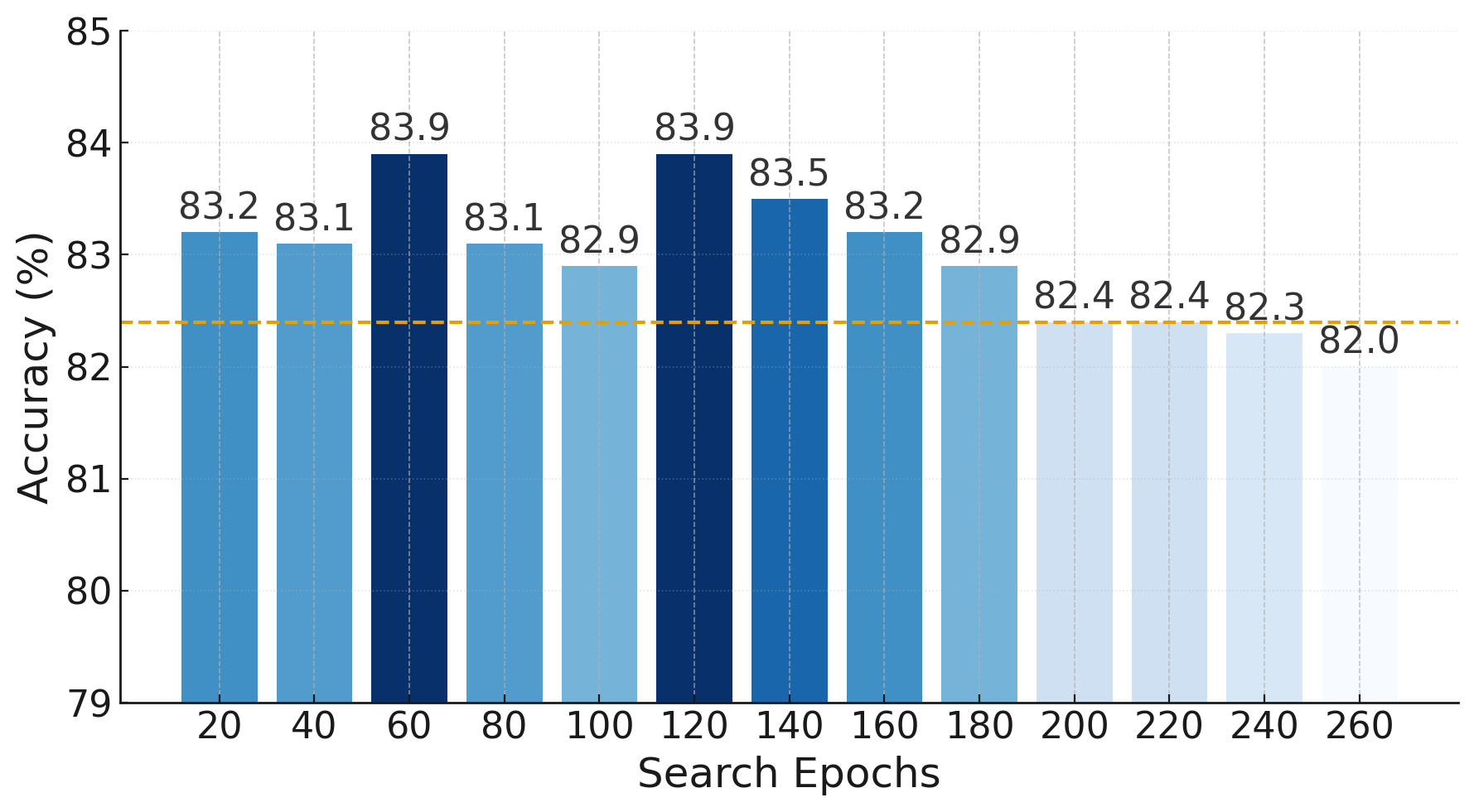}
    \end{tabular}
    \label{fig:ablaEP}
  \end{subfigure}\\[-2mm]
  % ----- (b) -----
  \hspace{-37mm}\begin{subfigure}[t]{0.48\textwidth}
    \centering
    \begin{tabular}{@{}m{0.01\textwidth} m{0.45\textwidth}@{}}
      \centering\hspace*{-0.5mm}(b) & \includegraphics[scale=0.125]{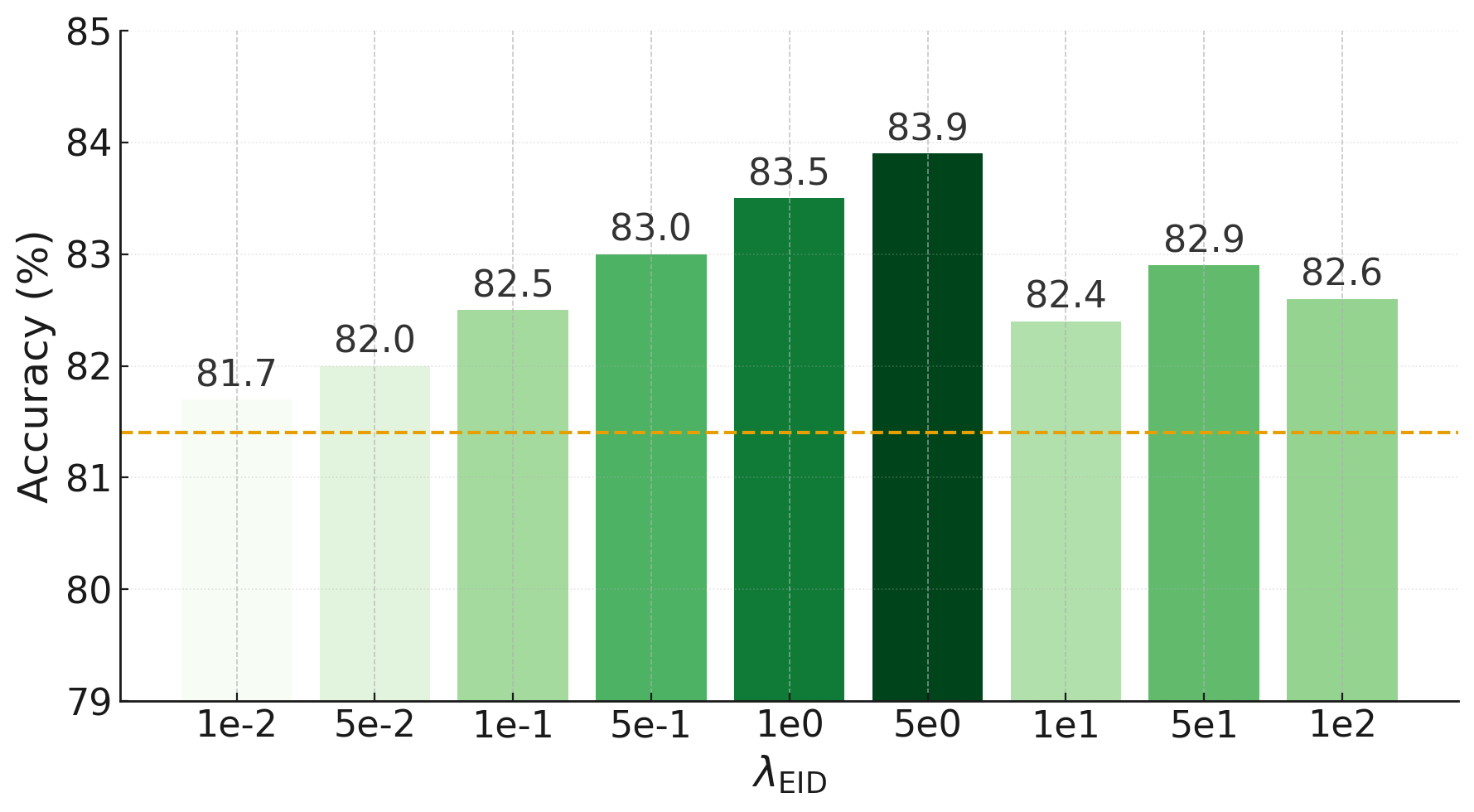}
    \end{tabular}
    \label{fig:ablaEID}
  \end{subfigure}
  \vspace{-4mm}
  \caption{Ablation with CIFAR10-DVS. (a) Ablation on different search epochs with a fixed total budget of 320 epochs (\(\text{search}+\text{finetune}=320\)). The dashed line marks the unpruned baseline (82.4\%). (b) Ablation on \(\lambda_{\mathrm{EID}}\). Bars show accuracies for non-zero \(\lambda\). The dashed line denotes \(\lambda=0\), where no EID is used.}
  \label{fig:abla_combo}
  \vspace{-4mm}
\end{figure}

\vspace{1mm}
\noindent\textbf{Ablation on Search epochs.}
Fig.~\ref{fig:abla_combo} (a) studies the split between search and finetune under a fixed total budget of 320 epochs on CIFAR10-DVS. We observe that the mask logits are learned effectively with \emph{few} search epochs: performance already exceeds the unpruned baseline at 20-40 search epochs, and peaks around \textbf{60-120} search epochs with accuracies up to \textbf{83.9\%}. Extending the search beyond 140 epochs brings diminishing returns and eventually degrades accuracy (\textit{e.g.,} 160-260 epochs: 83.2\% $\rightarrow$ 82.3\%), as the mask is obtained via probabilistic sampling and thus remains stochastic; it is important to \emph{freeze} a sampled mask at a suitable point and allocate \emph{sufficient} finetuning to consolidate it. An overlong search also squeezes the finetuning budget. These results indicate that with proper temperature annealing strategy, SpikeNM can quickly discovers effective \(N{:}M\) masks and benefits most from a modest search phase followed by adequate finetuning.

\vspace{1mm}
\noindent\textbf{Ablation on EID weight \(\lambda_{\text{EID}}\).}
Fig.~\ref{fig:abla_combo} (b) ablates the EID regularizer. 
Without EID (\(\lambda_{\text{EID}}=0\), dashed line), accuracy is 81.7\%. 
Small weights under-regularize (\textit{e.g.,} \(10^{-2}\!\sim\!10^{-1}\): 81.7-82.5\%), whereas moderate values steadily improve performance (from 83.0\% at \(5{\times}10^{-1}\) to a \textbf{peak of 83.9\%} at \(\lambda_{\text{EID}}{=}5\)). 
Pushing \(\lambda_{\text{EID}}\) too high begins to over-constrain the optimization and slightly harms accuracy (\textit{e.g.,} \(5\!\sim\!10^2\): 83.9-82.6\%). 
Overall, the curve demonstrates that EID \emph{consistently boosts} pruning stability and accuracy in a broad, practical range of \(\lambda_{\text{EID}}\), validating its effectiveness. Details of EID optimization are in Sec.~\ref{eid} of the appendix.

\vspace{1mm}
\noindent\textbf{Trade-off between accuracy and connectivity across pruning regimes.}
Table~\ref{tab:granu} provides a detailed comparison of the accuracy-connectivity trade-off.
At comparable levels of connectivity, \textbf{SpikeNM} consistently maintains a clear advantage.
Meanwhile, because UPR performs both weight and neuron pruning, we report two settings for fairness:
(i) our reimplementation of UPR that prunes weights only \emph{UPR*}; and
(ii) our method with \emph{additional neuron pruning} to match their setup.
With weight-only $N{:}M$ pruning, SpikeNM preserves almost the same accuracy as the unpruned model even under the $2{:}16$ pattern ($-0.1\%$) while using only 10.18\% connectivity.
When neuron pruning is further enabled, SpikeNM attains extremely low connectivity (6.93\%, 4.66\%, and 1.39\% for $2{:}4$, $2{:}8$, and $2{:}16$, respectively) and still achieves 82.5\% ($+0.1$\,pp), 81.7\% ($-0.7$\,pp), and 80.9\% ($-1.5$\,pp), outperforming the UPR baseline at similar connectivity.
These results shown the effectiveness and scalability of our approach across pruning regimes.

\vspace{1mm}
\noindent\textbf{Visualization.} For better understanding, we provide pruning visualizations in Sec. \ref{vis} of the appendix.

\begin{table}[t]
  \caption{Performance comparison between baselines on CIFAR10-DVS. ``$^\ast$” denotes our reimplementation.}
  \vspace{-3mm}
  \label{tab:granu}
  \centering
  \small
  \setlength{\tabcolsep}{4pt}
  \renewcommand{\arraystretch}{0.9}
  % 列：类别 / 方法 / 细粒度 / Acc / Δ / Conn.(%)
  \begin{tabular}{@{} l l l
                  S[table-format=2.1]
                  S[table-format=+2.1]
                  S[table-format=3.2] @{}}
    \toprule
    \textbf{Category} & \textbf{Method} & \textbf{Granularity} &
    \thU{Acc}{\%} &
    \thU{$\triangle$}{\%} &
    \thU{Conn.}{\%} \\
    \midrule
    
    % -------- Structured / SCA (2 rows)
    \multirow{2}{*}{Structured} & \multirow{2}{*}{SCA} & \multirow{2}{*}{Channel}
        & 72.8 & {+0.9} & 21.73 \\
    & & 
        & 71.9 & {-0.9} & 6.95 \\
    \cmidrule{1-6}
        
    % -------- Unstructured (12 rows total)
    \multirow{12}{*}{Unstructured} & \multirow{3}{*}{ESLSNN} & \multirow{3}{*}{Weight}
        & 80.6 & {-1.8} & 36.19 \\
    & & 
        & 79.2 & {-3.2} & 17.60 \\
    & & 
        & 77.5 & {-4.9} & 8.51 \\
    \cmidrule{2-6}
    & \multirow{3}{*}{STDS} & \multirow{3}{*}{Weight}
        & 81.7 & {-0.7} & 21.36 \\
    & & 
        & 81.1 & {-1.3} & 10.17 \\
    & & 
        & 79.8 & {-2.6} & 4.67 \\

    \cmidrule{2-6}
    & \multirow{3}{*}{UPR$^\ast$} & \multirow{3}{*}{Weight}
        & 82.1 & {-0.3} & 36.65 \\
    & & 
        & 81.6 & {-0.8} & 31.17 \\
    & & 
        & 80.0 & {-2.4} & 25.88 \\
        
    \cmidrule{2-6}
    & \multirow{3}{*}{UPR} & \multirow{3}{*}{\makecell[c]{Weight \& \\ Neuron}}
        & 81.9 & {-0.5} & 6.80 \\
    & & 
        & 81.0 & {-1.4} & 4.46 \\
    & & 
        & 79.0 & {-3.4} & 1.27 \\
    \cmidrule{1-6}
      
    % -------- Semi-structured (merged into one 6-row block)
    \multirow{6}{*}{\textbf{\makecell[c]{Semi-\\structured}}}
      & \textbf{2:4}  & \multirow{3}{*}{\textbf{Weight}}
        & \textbf{83.9} & \textbf{+1.5} & 43.32 \\
      & \textbf{2:8}  & 
        & \textbf{83.2} & \textbf{+0.8} & 24.41 \\
      & \textbf{2:16} &
        & \textbf{82.3} & \textbf{-0.1} & 10.18 \\
    \cmidrule{2-6}
      & \textbf{2:4}  & \multirow{3}{*}{\textbf{\makecell[c]{Weight \& \\ Neuron}}}
        & \textbf{82.5} & \textbf{+0.1} & 6.93 \\
      & \textbf{2:8}  & 
        & \textbf{81.7} & \textbf{-0.7} & 4.66 \\
      & \textbf{2:16} &
        & \textbf{80.9} & \textbf{-1.5} & 1.39 \\
    \bottomrule
  \end{tabular}
  \vspace{-2mm}
\end{table}

\section{Conclusion}
We presented \textbf{SpikeNM}, the first \(N{:}M\) semi-structured pruning framework for spiking neural networks. 
SpikeNM learns masks from scratch with an \(M\)-way basis logit parameterization and a differentiable sampler, which brings per block complexity down to \(\mathcal{O}(M)\) and yields sparsity patterns that are friendly to accelerator hardware. An eligibility inspired distillation regularizer further couples mask learning with spiking dynamics and improves stability under high sparsity.
On various datasets, SpikeNM with \(2{:}4\) and \(2{:}8\) sparsity preserves or improves accuracy compared with dense SNNs and consistently outperforms weight only pruning at matched connectivity. Combined with neuron pruning, it reaches very low connectivity while retaining a clear accuracy margin. In the future work, we will explore layer-wise adaptive \(N{:}M\), tighter weight and neuron pruning schedules, and deployment on sparse accelerators and neuromorphic hardware.

\clearpage
\newpage
{
    \small
    \bibliographystyle{ieeenat_fullname}
    \bibliography{main}
}


\end{document}